\documentclass{article}

\usepackage{arxiv}

\UseRawInputEncoding
\usepackage[T1]{fontenc}    % use 8-bit T1 fonts
\usepackage{hyperref}       % hyperlinks
\usepackage{url}            % simple URL typesetting
\usepackage{booktabs}       % professional-quality tables
\usepackage{amsfonts}       % blackboard math symbols
\usepackage{nicefrac}       % compact symbols for 1/2, etc.
\usepackage{microtype}      % microtypography
\usepackage{lipsum}
\usepackage{graphicx}

\title{

Weather impact on daily cases of COVID-19 in Saudi Arabia using machine learning}

\author{
  Abdullah Alsuhaibani \\
  Faculty of Computers and Information Systems\\
  Islamic University of Madinah\\
   Madinah, Saudi Arabia \\
  \texttt{abdullah.alsuhaibani@iu.edu.sa} \\
   \And
 Abdulrahman Alhaidari \\
  Faculty of Computers and Information Systems\\
  Islamic University of Madinah\\
   Madinah, Saudi Arabia \\
  \texttt{a.alhaidari@iu.edu.sa} \\
}

\begin{document}
\maketitle

\begin{abstract}
COVID-19 was announced by the World Health Organisation (WHO) as a global pandemic.  The severity of the disease spread is determined by various factors such as the countries' health care capacity and the enforced lockdown. However, it is not clear if a country's climate acts as a contributing factor towards the number of infected cases. This paper aims to examine the relationship between COVID-19 and the weather of 89 cities in Saudi Arabia using machine learning techniques. We compiled and preprocessed data using the official daily report of the Ministry of Health of Saudi Arabia for COVID-19 cases and obtained historical weather data aligned with the reported case daily reports. We preprocessed and prepared the data to be used in models' training and evaluation. Our results show that temperature and wind have the strongest association with the spread of the pandemic. Our main contribution is data collection, preprocessing, and prediction of daily cases.
 For all tested models, we used cross-validation of K-fold of K=5. Our best model is the random forest that has a Mean Square Error(MSE), Root Mean Square (RMSE), Mean Absolute Error (MAE), and R{2} of 97.30, 9.86, 1.85, and 82.3\%, respectively.         
\end{abstract}

% keywords can be removed


\section{Introduction}
Last year, 2020, is considered to be one of the most challenging years because of the consequences of the disease that started in Wuhan, China, on December 31, 2019. The pandemic hit many countries, and many families lost their beloved ones, and companies around the world have suffered from its consequences. The World Health Organisation (WHO) reported that confirmed cases reached more than 87 million and nearly 2 million deaths on January 10, 2021.\cite{WHOCoron60:online} It is viewed as a large-scale pandemic that invaded and disrupted all people's activities. The number of cases is still increasing, till the time of writing this paper, on a daily basis, and the mortality rate is rising substantially. 

Fever, dry cough, and fatigue are the primary symptoms of COVID-19. Nevertheless, some patients who were diagnosed with COVID-19 show no symptoms, which can expedite the disease infection rate .\cite{doi:10.1177/1756286420917830}. The study in \cite{zhou2020clinical} shows the correlation between elderly patients and death cases for COVID-19.

On March 2, 2020, the first case was diagnosed and announced by the Saudi Ministry of Health that is located in east providence Qatif City. Saudi Arabia's government has taken some serious actions to prevent the spread of the virus, including the shutdown of borders, schools, the complete ban between regions as well as Umrah suspension \cite{anil2021impact}. Saudi Arabia is located in southwestern Asia that is closer to Africa. According to the General Authority for Statistics (GASTAT), the total area the country occupied is 2 million square kilometers with a population of 33.413.660 \cite{GeneralI55:online} \cite{Statisti8:online}. Saudi Arabia is considered one of the largest countries in the Middle East. Riyadh is the capital of Saudi Arabia, with a total population of more than 8 million, whereas Makkah and Medina combined have a total population of approximately 10.3 million people, and both are the holy places in the Kingdom of Saudi Arabia \cite{Emirate:online}. 

The weather in Saudi Arabia is mild to hot in all seasons except in some regions where the winter is cold. The goal of this study is to correlate the weather and the spread of COVID-19 in Saudi Arabia. We have applied supervised machine learning algorithms to analyze the behavioral effect of weather on the disease. The aim is to predict the daily cases of COVID-19 in Saudi Arabia, given the climate of different cities. Selecting the best state-of-the-art machine learning models and features impacts on COVID-19 spreading in Saudi Arabia is explored in this paper.

\section{Related Work}
In the work of \cite{bukhari2020effects}, the authors examined the relationship between the weather in Boston, United States, and COVID-19.  The result indicated that 85\% out of the total number of reported cases were recorded in places where the weather is between 3C and 17C. Seven variables were used by \cite{mofijur2020relationship} in Dhaka, Bangladesh, to correlate the climate and the spread of the pandemic. The authors concluded that only two attributes of the collected data were significant in the analysis, minimum temperature and average temperature. In another study in Canada, the authors used statistical analysis to draw a relationship between the COVID-19 cases and the climate. The result showed that there was no association between high temperature and the increase of the newly recorded cases\cite{to2020correlation}. 

 Time series prediction was used in the prediction of \cite{notari2020temperature} in different countries. They use a starting point of 30 cases, and then they fitted the consecutive 12 days. They foretasted that the northern hemisphere countries on the globe would have as a result of hot weather and lockdown policies. 

A study by \cite{demongeot2020temperature} investigated 21 different countries in addition to the regions administrated by French. They used publicly available data and concluded that as the temperature decreases, the COVID-19 cases do the same.

 The authors in \cite{malki2020association} conclude that weather features such as temperature and humidity predicted death rate as well as confirmed cases. KNN and Decision Tree obtained the highest performance models to predict confirmed cases and death cases. In\cite{rendana2020impact} shows that a low wind speed contributes to the increase of the number of infected cases for COVID-19 in Jakarta, Indonesia, by using the Spearman correlation test. The paper in \cite{yadav2020analysis} proposed a support vector regression model to predict a COVID-19 prevalence across countries including (Mainland China, US, Italy, South Korea and, India) pandemic comes to an end, and analyses the growth and transmission rates. Using Pearson's correlation, humidity and temperature were important factors for increasing positive cases in both New York and Milan. In Jakarta, Indonesia \cite{tosepu2020correlation} determined that weather particularly temperature average has correlation with spread of COVID-19 cases where 9999 (r = 0.392; p < .01)by using Spearman correlation coefficient.
 
The study in \cite{babekercorrelation} shows the relation between COVID 19 daily cases and weather features in two major cities Riyadh and Makkah. The period for this study was 35 days starts from 8 April 2020 until 13 May 2020. The high temperature results in high daily cases when the temperature between 19 C to 42 C. However, the study was applied to two cities in Saudi Arabia.
Long short-term memory (LSTM), a deep learning approach, was trained by \cite{elsheikh2020deep} to predict total new cases, cured cases, and mortality in Saudi Arabia utilizing official data by the Saudi ministry of health.  The result was measured by seven metrics, such as Mean square error (MSE). The model was used to predict the new cases of COVID-19 in six different countries \cite{elsheikh2020deep}.

\section{Methodology}
\subsection{Data Collection}
\begin{table}[t]
 \caption{Daily Cases }
  \centering
  \begin{tabular}{lllll}
    \toprule
    \cmidrule(r){1-2}
    City     & Daily Cases &Daily Moralities \\
    \midrule

Makkah & 3987 & 58\\
Riyadh	& 2980 &6\\
Medina	& 2830&32\\
Jeddah	& 2738&28\\
Dammam	& 957&1\\
Hufof	& 925&3\\
Taif	& 241&0\\
Jubail	& 211&1\\
Qatif	& 200&1\\
Tabuk	& 185&1\\

    \bottomrule
  \end{tabular}
  \label{tab:Daily}
\end{table}

\subsubsection{COVID-19 Dataset}

The dataset was gathered from the Ministry of Health of the Kingdom of Saudi Arabia's official website\cite{COVID19D0:online}. According to the General Authority for Statistics (GASTAT), the total area the country occupied is 2 million square kilometers with a population of 33.413.660 \cite{GeneralI55:online} \cite{Statisti8:online}. At the time where the data was collected for this study, the dataset included 89 cities around Saudi Arabia divided among 13 regions that COVID-19 infected. 
The data from March 2020 until April 2020 were present in this study. The total number of records is 4860, where features as follows: date, city name, region name, cumulative cases, cumulative recoveries, cumulative moralities, cumulative active cases, daily cases, daily recoveries, and daily moralities. As shown in table \ref{tab:Daily}, the total number of reported cases is 15254 based on the daily infections for the top 10 cities is nearly 93\% of the total daily cases. Mortality in Makkah was the highest.

 \subsubsection{Weather Dataset}
 
 The data were obtained from the Open Weather website using their API https://openweathermap.org/. The retrieval information of weather is based on 13 regions along with Jeddah and Taif. Population density and climatic weather lead us to handle them as regions. The start date was on the same period of the study and came with 26 features. However, the needed features were the following: feels like, humidity, pressure, temperature, temperature by hours, maximum temperature, minimum temperature, weather condition, wind degree, wind speed, and weather main. Other features were ignored because either empty values such as sea level, snow, and rain or ineffective for our experiments like timezone and ID.

\subsection{Methods}
In order to establish a relationship between the pandemic and the daily new cases, we used different machine learning algorithms. These were trained to detect the pattern and the trend by machine learning models. We used python and Sklearn that allow us to train the preprocessed data and used these models on unseen, new data. Cross-validation was utilized, meaning that our data is divided into training, validation, and testing. Then, the best subset of the data is used for training the model, based on the performance of the validation. Next, the model is tested on separate testing data. The utilized algorithms are summarized in the result section.
For evaluation, each model was trained separately, and the main goal is to select the best performing model on the collected data. Aiming to increase the models' ability to detect daily case changes as the weather changing, predicting daily cases was feasible. Since the target variable we are looking at is regression, we focused on the Mean Square Error (MSE), which is the primary metric for evaluation. 

For the prediction, the used features are in the table \ref{tab:features}

\begin{table}[ht]
 \caption{Features used in the models}
  \centering
  \begin{tabular}{lllll}
    \toprule
    Feature Name & Excluded?& Feature Name & Excluded?   \\
\midrule
Cumulative active cases& Yes&Temperature by hours& No\\
Cumulative cases& No & Temperature maximum&	No\\
Cumulative moralities& No & Temperature minmium&	No\\
Cumulative recoveries& No&	Weather main (i.e. rain, or cloudy) & Yes\\	
Daily moralities& No &Wind degree&Yes	\\
Daily recoveries&No	&Wind speed&Yes\\
Feels like&	Yes& City & Yes\\
Humidity& Yes	& Dates& No\\
Pressure&	Yes\\
Temperature& Yes\\
    \bottomrule
  \end{tabular}
  \label{tab:features}
\end{table}

For using features, it is determined by one of two possibilities. First, if the feature does not significantly impact the prediction, or we thought it would leakage data, we exclude it from the projection. The latter is tremendously essential because if the model received a hint about the unseen data, it can not be generalized and may fail to completely new predictions out of the testing data.  

In order to discover the most important features that affect the response, Relief, which is an algorithm that measures the sensitive features to response, was used. 
Also, We used different machine learning algorithms to associated the COVID-19 pandemic with the weather. After data prepossessing, multiple machine learning models were created. We based our work on the classical machine learning algorithms, as the main goal is finding a relation between COVID-19 spread and weather fluctuations. We include not only temperature but also other weather's properties such as wind speed and degree. The collected data has 16 features for each day in 89 regions of Saudi Arabia as time series. The target for the created models is the daily cases. However, to prevent data leakage, we removed some features to prevent the model from inferring the number of newly infected cases from an existing feature. 
In our approach, we use time-series as we have each date's infection data. Then, the data items were sorted based on the dates range sequence. Subsequently, the dates were removed after sorting. The reason is that we encoded the dates using one-hot encoding. For instance, if a case was reported in Madinah on March 20, 2020, then only this date will have a value of one, and the proceeding and the succeeding dates will have zeros. Therefore, numerous columns were created. This led to not only minimization of the training time but also a slight improvement of the model performance was sought.

It turned out that not all features are essential when it comes to the final response. For instance, when both daily cases and mortality were included, the model's performance degraded substantially, as we use ranked and sorted features, they have outlined in the table \ref{tab:features}. Additionally, cumulative deaths were excluded because we do not want the model to have information related to daily cases, our prediction.  

\section{Results and Discussion}

This section summarises our findings and contributions made. Our findings on COVID-19 and its relation to the disease spread at least hint that as the weather changes, there is a change in the number of daily cases. In the table \ref{tab:Results} we report the models that were trained on the weather and COVID-19 data.

\begin{table}[ht]
 \caption{Models' evaluation}
  \centering
  \begin{tabular}{lllll}
    \toprule
    Model     & MSE     & RMSE & MAE & $R^{2}$  \\
    \midrule 

 Random forest & 97.30&9.86&1.85 & 82.3\% \\
AdaBoost& 121.65&11.03&1.95& 76.7\% \\
 Decision tree&167.14&12.92&2.38& 69.2 \%\\
KNN &249.21&15.78&2.80& 54.3 \%\\
Neural Network&356.88&18.89&4.41&34.5 \%\\

    \bottomrule
  \end{tabular}
  \label{tab:Results}
\end{table}

The results show that the best performing model among all is random forest. It is vivid that the neural network was not able to capture the variability of the training data rather than testing samples. Therefore, it overfit the training; consequently, it has a mediocre performance on the testing set. 
All features were used in the final prediction except the dates of when the cases were reported. This was determined to reduce the training time, and more importantly, the results of the experiments were not significantly affected by removing the dates. We ranked the features according to their influence on the result. The most significant was wind degree followed by temperature. 

\begin{figure}[ht]
\label{fig:tmp}
\includegraphics[width=6cm]{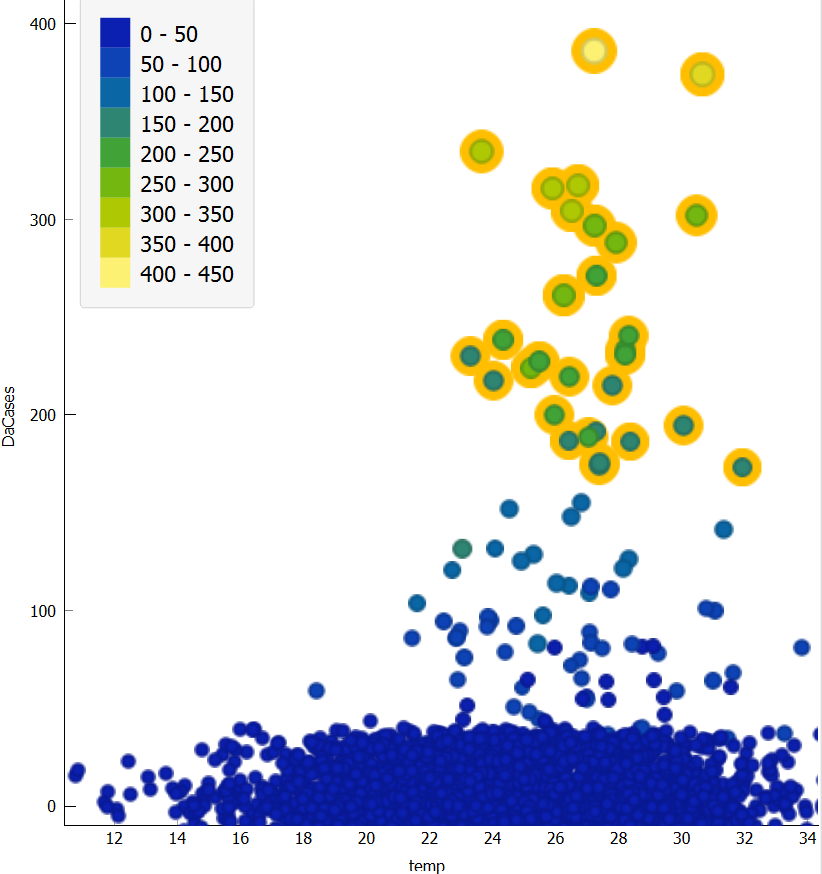}
\centering
\caption{Temperature and daily cases correlation }
\end{figure}

In figure \ref{fig:tmp}, when the temperature is between 25 and 32 C, we see from the data that the daily cases increase. When it reached 32.5 C, it starts to decrease. Moreover, we see that the more the heat increase does not necessarily affect the reported case.  Weather main such as raining is effecting the cases after the temperature. 

One concern about the findings was that the data used is from the beginning of the COVID-19 spread in Saudi Arabia, where there are fewer cases than in other countries. It remains unclear that in the extreme cold climate, which is not the case of Saudi Arabia, to which degree weather is attributed to increasing or decreasing in COVID-19 daily cases.

\section{Conclusion}

The findings of this study can be understood as there is a correlation between weather and the infection of COVID-19. Also, our experiments concluded that the most critical weather feature is temperature, wind speed, feels like, wind degree and humidity. For future research, we suggest using different AI techniques such as deep learning with more diverse data from other parts of the world. Besides, we recommend that using COVID-19 and weather data of a country where the pandemic has been around for several months, and the number of cases is significantly high. This likely makes the correlation clearer as the number of daily cases may significantly change with the weather fluctuation.



%%% \cite{jensensvendborg}

%\bibliographystyle{unsrt}  
%\bibliography{main}  %%% Remove comment to use the external .bib file (using bibtex).

\bibliographystyle{unsrt}
\bibliography{references}

\begin{thebibliography}{10}

\bibitem{WHOCoron60:online}
Who coronavirus disease (covid-19) dashboard | who coronavirus disease
  (covid-19) dashboard.
\newblock \url{https://covid19.who.int/}.
\newblock (Accessed on 01/26/2021).

\bibitem{doi:10.1177/1756286420917830}
Hai-Yang Wang, Xue-Lin Li, Zhong-Rui Yan, Xiao-Pei Sun, Jie Han, and Bing-Wei
  Zhang.
\newblock Potential neurological symptoms of covid-19.
\newblock {\em Therapeutic Advances in Neurological Disorders},
  13:1756286420917830, 2020.
\newblock PMID: 32284735.

\bibitem{zhou2020clinical}
Fei Zhou, Ting Yu, Ronghui Du, Guohui Fan, Ying Liu, Zhibo Liu, Jie Xiang,
  Yeming Wang, Bin Song, Xiaoying Gu, et~al.
\newblock Clinical course and risk factors for mortality of adult inpatients
  with covid-19 in wuhan, china: a retrospective cohort study.
\newblock {\em The lancet}, 395(10229):1054--1062, 2020.

\bibitem{anil2021impact}
Ismail Anil and Omar Alagha.
\newblock The impact of covid-19 lockdown on the air quality of eastern
  province, saudi arabia.
\newblock {\em Air Quality, Atmosphere \& Health}, 14(1):117--128, 2021.

\bibitem{GeneralI55:online}
General information about the kingdom of saudi arabia | general authority for
  statistics.
\newblock \url{https://www.stats.gov.sa/en/4025}.
\newblock (Accessed on 02/14/2021).

\bibitem{Statisti8:online}
Statistical yearbook of 2019 | issue number: 55 | general authority for
  statistics.
\newblock \url{https://www.stats.gov.sa/en/1006}.
\newblock (Accessed on 02/14/2021).

\bibitem{Emirate:online}
Emirate of~Makkah~region.
\newblock \url{https://www.makkah.gov.sa/en}.
\newblock (Accessed on 05/2/2021).

\bibitem{bukhari2020effects}
Qasim Bukhari, Joseph~M Massaro, Ralph~B D’agostino, and Sheraz Khan.
\newblock Effects of weather on coronavirus pandemic.
\newblock {\em International journal of environmental research and public
  health}, 17(15):5399, 2020.

\bibitem{mofijur2020relationship}
M~Mofijur, IM~Rizwanul~Fattah, ABM Saiful~Islam, MN~Uddin, SM~Ashrafur~Rahman,
  MA~Chowdhury, Md~Asraful Alam, Md~Uddin, et~al.
\newblock Relationship between weather variables and new daily covid-19 cases
  in dhaka, bangladesh.
\newblock {\em Sustainability}, 12(20):8319, 2020.

\bibitem{to2020correlation}
Teresa To, Kimball Zhang, Bryan Maguire, Emilie Terebessy, Ivy Fong, Supriya
  Parikh, and Jingqin Zhu.
\newblock Correlation of ambient temperature and covid-19 incidence in canada.
\newblock {\em Science of the Total Environment}, 750:141484, 2020.

\bibitem{notari2020temperature}
Alessio Notari.
\newblock Temperature dependence of covid-19 transmission.
\newblock {\em arXiv preprint arXiv:2003.12417}, 2020.

\bibitem{demongeot2020temperature}
Jacques Demongeot, Yannis Flet-Berliac, and Herv{\'e} Seligmann.
\newblock Temperature decreases spread parameters of the new covid-19 case
  dynamics.
\newblock {\em Biology}, 9(5):94, 2020.

\bibitem{malki2020association}
Zohair Malki, El-Sayed Atlam, Aboul~Ella Hassanien, Guesh Dagnew, Mostafa~A
  Elhosseini, and Ibrahim Gad.
\newblock Association between weather data and covid-19 pandemic predicting
  mortality rate: Machine learning approaches.
\newblock {\em Chaos, Solitons \& Fractals}, 138:110137, 2020.

\bibitem{rendana2020impact}
Muhammad Rendana.
\newblock Impact of the wind conditions on covid-19 pandemic: A new insight for
  direction of the spread of the virus.
\newblock {\em Urban climate}, 34:100680, 2020.

\bibitem{yadav2020analysis}
Milind Yadav, Murukessan Perumal, and M~Srinivas.
\newblock Analysis on novel coronavirus (covid-19) using machine learning
  methods.
\newblock {\em Chaos, Solitons \& Fractals}, 139:110050, 2020.

\bibitem{tosepu2020correlation}
Ramadhan Tosepu, Joko Gunawan, Devi~Savitri Effendy, Hariati Lestari, Hartati
  Bahar, Pitrah Asfian, et~al.
\newblock Correlation between weather and covid-19 pandemic in jakarta,
  indonesia.
\newblock {\em Science of The Total Environment}, page 138436, 2020.

\bibitem{babekercorrelation}
EA~Babeker.
\newblock Correlation between some climatic factors and covid-19 epidemic in
  two cities in kingdom of saudi arabia.

\bibitem{elsheikh2020deep}
Ammar~H Elsheikh, Amal~I Saba, Mohamed Abd~Elaziz, Songfeng Lu, S~Shanmugan,
  T~Muthuramalingam, Ravinder Kumar, Ahmed~O Mosleh, FA~Essa, and Taher~A
  Shehabeldeen.
\newblock Deep learning-based forecasting model for covid-19 outbreak in saudi
  arabia.
\newblock {\em Process Safety and Environmental Protection}, 149:223--233,
  2020.

\bibitem{COVID19D0:online}
Covid 19 dashboard: Saudi arabia.
\newblock \url{https://covid19.moh.gov.sa/}.
\newblock (Accessed on 02/14/2021).

\end{thebibliography}


\end{document}